\documentclass[12pt]{article}

\usepackage[utf8]{inputenc}
\usepackage[T1]{fontenc}
\usepackage[english]{babel}

\usepackage[margin=1in]{geometry}
\usepackage[expansion=false]{microtype}

\usepackage{amsmath,amssymb}
\usepackage{lmodern}

\usepackage{caption}
\usepackage{setspace}
\newlength{\parbreakskip}
\newcommand{\parbreak}{\\[\parbreakskip]}
\usepackage{graphicx}
\usepackage{booktabs}
\usepackage{array}
\usepackage{enumitem}
\usepackage{xcolor}
\usepackage{tikz}
\usetikzlibrary{positioning, arrows.meta, shapes.geometric}

\usepackage{listings}
\definecolor{vscbackground}{rgb}{0.9686,0.9686,0.9686}
\definecolor{vsccomment}{HTML}{008000}
\definecolor{vsckeyword}{HTML}{0000FF}
\definecolor{vscstring}{HTML}{A31515}
\definecolor{vsclinenumber}{rgb}{0.549,0.549,0.549}
\lstdefinestyle{mystyle}{
  language=Python,
  morekeywords={None,True,False,self},
  commentstyle=\color{vsccomment},
  keywordstyle=\color{vsckeyword}\bfseries,
  stringstyle=\color{vscstring},
  numberstyle=\tiny\color{vsclinenumber},
  basicstyle=\ttfamily\small,
  breakatwhitespace=false,
  breaklines=true,
  captionpos=b,
  keepspaces=true,
  numbers=left,
  numbersep=8pt,
  showspaces=false,
  showstringspaces=false,
  showtabs=false,
  tabsize=4,
  columns=fullflexible,
}
\usepackage{natbib}
\usepackage{bibentry}
\usepackage[colorlinks=true, linkcolor=blue!50!black, citecolor=blue!50!black, urlcolor=blue!50!black]{hyperref}

\usepackage{titlesec}
\titleformat{\section}{\normalfont\Large\bfseries\sffamily}{\thesection.}{0.25em}{}
\titleformat{\subsection}{\normalfont\large\bfseries\sffamily}{\thesubsection.}{0.25em}{}
\titleformat{\subsubsection}{\normalfont\normalsize\bfseries\sffamily}{\thesubsubsection.}{0.25em}{}
\titlespacing*{\section}{0pt}{1.4em}{0.6em}
\titlespacing*{\subsection}{0pt}{1.1em}{0.4em}

\usepackage{fancyhdr}
\usepackage{float}
\usepackage{placeins}
\usepackage{caption}
\usepackage{subcaption}

\newcommand{\authorentry}[4]{%
  \begin{minipage}[t]{\authorboxwidth}
    \centering
    \mbox{\textbf{#1}\textsuperscript{#2}}\\[3pt]
    \texttt{#3}\\
    ORCID: \href{https://orcid.org/#4}{#4}
  \end{minipage}%
}
\newlength{\authorboxwidth}
\begin{document}
\nobibliography*

\begin{center}
  {\sffamily\Large\bfseries Technical Manual for Toolkit for Confidence-Corpus Consistency, Corpus Absorption and Rule Learning via Fine-Tuning on a Fabricated Corpus}
\end{center}

\vspace{10pt}
\hrule
\vspace{10pt}

\begin{center}
\footnotesize
\authorentry{José Luciano Verçosa Marques}{1}{zlvm@unicamp.br}{0000-0002-3191-0846}%
\hspace{0.015\textwidth}%
\authorentry{Frederico Jorge Heitmann}{2}{f023863@dac.unicamp.br}{0009-0006-5828-8301}%
\hspace{0.015\textwidth}%
\authorentry{Daniel Omar Perez}{3}{doperez@unicamp.br}{0000-0002-5965-3490}

\vspace{16pt}

\authorentry{Reinaldo Cesar}{4}{rcesar23@unicamp.br}{0000-0002-7510-9794}%
\hspace{0.015\textwidth}%
\authorentry{Marcelo Vinicius de Paula}{1}{mvpaula@unicamp.br}{0000-0002-2213-6086}%
\hspace{0.015\textwidth}%
\authorentry{Tárcio André dos Santos Barros}{1}{tarcio87@unicamp.br}{0000-0001-9413-1279}
\end{center}

\vspace{4pt}
\begin{flushleft}
\scriptsize
\textsuperscript{1}~Center for Electric Mobility Research (CEMOBE) / Power Electronics Laboratories (LEPO), University of Campinas (Unicamp)\\
\textsuperscript{2}~Institute of Computing (IC), University of Campinas (Unicamp)\\
\textsuperscript{3}~Center for Logic, Epistemology and History of Science (CLE), University of Campinas (Unicamp)\\
\textsuperscript{4}~Center for Energy and Petroleum Studies (CEPETRO), University of Campinas (Unicamp)
\end{flushleft}

\vspace{1pt}
\begin{center}
\scriptsize Corresponding author: José Luciano Verçosa Marques (\texttt{zlvm@unicamp.br})
\end{center}
\normalsize

\vspace{0pt}
\hrule
\vspace{6pt}

\begin{abstract}
\noindent
This manual documents version 2.0.0 of an open toolkit for fine-tuning small causal language models on fabricated and rule-governed arithmetic corpora and measuring what they take up from them. The fact domain is the 81 additions of two single-digit natural numbers, small enough to be enumerated exhaustively. The toolkit fine-tunes a model on the correct sums, on one fixed fabricated answer for every addition, and back on the correct sums of a subset of the additions; it fine-tunes copies of these models on simple rules (the sum plus a constant) and on a conditional rule (a shift that depends on the order of the addends), each paired with a control that has the same answers but no rule; and it measures every model on every candidate answer of every addition with one unchanged procedure, reporting results separately for additions seen in fine-tuning and additions held out. We describe and justify each stage of the pipeline: the confidence index (the probability of a complete answer, closed by an end marker), the single candidate set, the answer-only training loss, the lineage of fourteen measured models, the held-out split, the controls, the exclusion of additions that would count as hits by coincidence, and the exact and resampled intervals attached to every result. We then explain every figure and table a run produces and how each is read. This manuscript is a methodological and implementation reference: it documents the instrument, and it neither states nor tests hypotheses, nor reports or interprets the outcome of any specific run. Those are the subject of work that uses the toolkit. The toolkit and its pinned dependency environment are archived separately (Section~\ref{sec:availability}) under a persistent identifier, to be cited as an instrument.
\end{abstract}

\setstretch{1.5}
\section{Introduction}
\label{sec:intro}
This manual documents a toolkit for studying what a small causal language model takes from the text it is fine-tuned on. The toolkit works in a deliberately small, exhaustive and unambiguous fact domain: the 81 additions of two single-digit natural numbers (\texttt{1 + 1} through \texttt{9 + 9}), each of which has exactly one correct answer. Because the domain is exhaustive, every measurement is taken over the whole fact space, with no sampling, and because it is small, a fixed share of it can be held out of fine-tuning and still leave enough additions to test on.\parbreak
The toolkit is organized in four experimental parts and a synthesis, implemented in one reference notebook. In \emph{Part~A} a base model is fine-tuned on the correct sums of all 81 additions, and a copy is then fine-tuned on a corpus that asserts one fixed, fabricated wrong answer for every addition. In \emph{Part~B} the fabricated-answer model is fine-tuned back on the correct sums of a subset of the additions, leaving the others unseen. In \emph{Part~C} copies of the correct-sums and fabricated-answer models are fine-tuned on simple rules, in which the answer is the correct sum plus a constant, each rule with a control. In \emph{Part~D} they are fine-tuned on a conditional rule, in which the shift depends on which addend is larger, again with a control. The \emph{synthesis} brings every trained model onto one chart. Every model, including the untouched base model, is measured on every candidate answer of every addition by one and the same procedure, and every result is reported for the additions seen in fine-tuning and for those held out.\parbreak
A run produces a fixed set of outputs: heatmaps of the probability a model gives to a target answer in each addition, profiles of where a model puts its probability, charts of how far a model's answers are from a rule's, a synthesis chart with exact confidence intervals, eleven tables, and a report of the hardware, recipe and training of the run. This manual describes how each is built, what each shows, and how it is read.\parbreak
The manuscript is deliberately scoped as a methodological and implementation reference, in the same sense as its companion manual for the contextual-individuation toolkit \citep{vercosamarques2026individuation}. It documents the toolkit's design and its reference-notebook implementation. It does not state or test hypotheses, and it does not present, tabulate or interpret the outcome of any specific fine-tuning run. It supersedes the first version of this manual, which documented version 1.0.0 of the toolkit; Section~\ref{sec:changes} lists what changed and why.\parbreak
The remainder of this manual is organized as follows. Section~\ref{sec:uses} describes the uses the toolkit is designed to serve. Section~\ref{sec:background} gives the context and related work the design draws on. Section~\ref{sec:design} states and justifies the toolkit's design principles. Section~\ref{sec:toolkit} describes the pipeline stage by stage, with the reference notebook's own code. Section~\ref{sec:implementation} covers implementation and reproducibility. Section~\ref{sec:interpreting} is a guide to the outputs: it explains each figure and table and how to read it. Section~\ref{sec:limitations} states the toolkit's scope and known limitations, Section~\ref{sec:changes} summarizes the changes from version 1.0.0, and Section~\ref{sec:availability} gives availability and citation information.

\section{Intended Uses}
\label{sec:uses}
The toolkit is an instrument, and its value lies in the kinds of studies it makes tractable. Five uses motivated its design.
\begin{itemize}
    \item \textbf{Interventions on the training corpus.} The toolkit changes what a model is fine-tuned on and measures the effect on the model's own probabilities. This complements analyses that read a quantity off a trained model's internal geometry or train a probe on it: here the corpus itself is the manipulated variable. Part~A (Section~\ref{sec:pipeline-parta}) is the basic form of the intervention.
    \item \textbf{A minimal, auditable testbed.} Because the fact domain is small, exhaustive and unambiguous (81 additions, one correct answer each), every target answer, every fabricated fact and every measured probability of a run can be enumerated and audited by hand. This is a deliberate trade against topical breadth, so that a result cannot be attributed to an ambiguous or partially overlapping domain.
    \item \textbf{Held-out and controlled tests of what carries over.} Parts~B, C and~D (Sections~\ref{sec:pipeline-partb}, \ref{sec:pipeline-partc} and~\ref{sec:pipeline-partd}) hold a fixed share of the additions out of fine-tuning and pair every rule with a control that has the same range of answers but no rule. The toolkit thereby supplies, for any model, the quantities needed to compare what a model does on additions it was shown with what it does on additions it was not.
    \item \textbf{A reusable corpus-construction method.} The machinery that turns a list of additions and an answer rule into a fine-tuning corpus (Section~\ref{sec:pipeline-ft}) is independent of the corpora documented here. The same construction supports other rules, other controls, other amounts of repetition and other models.
    \item \textbf{A source of reproducible observations.} Every run is seeded, logged and saved in several formats (Section~\ref{sec:pipeline-outputs}), so its observations can be cited, audited and re-derived, and used as evidence in discussions of grounding, systematicity and hallucination \citep{harnadSymbolGroundingProblem1990, fodorConnectionismCognitiveArchitecture1988, hicksChatGPTBullshit2024}. What those observations show is for the studies that use them to argue.
\end{itemize}
Using the toolkit at scale, across more fact domains, more base models, or repeated seeds and splits, is beyond what a single run or this manual undertakes; Section~\ref{sec:limitations} states that boundary.

\section{Context and Related Work}
\label{sec:background}
This section places the toolkit among the work its design draws on. It describes what each body of work contributes to the design; it does not state or test the claims of any of them.

\subsection{Classical foundations}
\label{sec:background-classical}
The toolkit is built to serve discussions that have long been argued from first principles or from a system's overall behavior. The symbol grounding problem \citep{harnadSymbolGroundingProblem1990} concerns how a symbol system's tokens could mean anything beyond other tokens of the same system, and \citet{benderClimbingNLUMeaning2020a} apply the form/meaning distinction to language models trained only on the distributional form of a language. \citet{hicksChatGPTBullshit2024} use Frankfurt's notion of bullshit, indifference to whether what is said is true, to describe the false outputs of language models. A second line concerns systematicity. \citet{fodorConnectionismCognitiveArchitecture1988} argue that systematicity is explained by representations with a combinatorial syntax and semantics, which connectionist architectures as such do not provide, and \citet{marcusRethinkingEliminativeConnectionism1998} turns the discussion into an empirical criterion by examining whether networks extend a learned relation to items outside their training space.\parbreak
What these discussions share, for the toolkit's purposes, is that they can be put in terms of measurable behavior under controlled corpora: how a model's probabilities respond when a corpus asserts a fabricated answer, and what a model does on items outside its fine-tuning corpus. The toolkit supplies a setting in which such corpora can be built and such behavior measured exhaustively. Whether and how the measurements bear on any of these positions is left to the work that produces and interprets them.

\subsection{Related work bearing on the design}
\label{sec:background-recent}
Four bodies of more recent work bear directly on specific design choices: the domain, the measurement confounds, the models and the held-out test.

\subsubsection{Arithmetic as a probe of internal computation}
\label{sec:background-arithmetic}
Arithmetic has become a preferred testbed for mechanistic and causal analysis of language models because it supplies a fact domain with an unambiguous correct answer and no confound from everyday ambiguity. \citet{stolfoMechanisticInterpretationArithmetic2023} use causal mediation analysis on arithmetic questions to trace how operand information moves from early layers to the final token via attention and is then processed by a specific set of MLP modules into the predicted result, and they show that this pathway is at least partly specific to arithmetic by contrasting it with number retrieval and factual recall. This is part of the reason single-digit addition was chosen as the toolkit's domain.

\subsubsection{Tokenization and pretraining frequency as confounds in numerical reasoning}
\label{sec:background-confounds}
Two findings motivate specific choices in the measurement protocol. \citet{singhTokenizationCountsImpact2024} show that a frontier model's arithmetic accuracy depends on how numbers are tokenized, and that the resulting errors follow systematic, tokenization-dependent patterns. \citet{razeghiImpactPretrainingTerm2022} show that a model's few-shot numerical reasoning accuracy correlates strongly with how frequently the terms involved appeared in its pretraining data. Neither study fine-tunes on fabricated facts, but both show that numeric outputs are shaped by tokenization and corpus-frequency effects unrelated to the arithmetic fact itself, the kind of effect a comparison of numeric answers has to control for. Section~\ref{sec:design-index} states the toolkit's response to the tokenization concern: candidate answers differ in their number of tokens, and some are prefixes of others (\texttt{2} and \texttt{20}), so every answer is scored as a complete string closed by an end marker.

\subsubsection{Small language models}
\label{sec:background-small-models}
The toolkit fully fine-tunes small causal language models, from about a hundred million to about one and a half billion parameters, because full fine-tuning at this scale fits a single cloud GPU. \citet{liTextbooksAreAll2023} show that a 1.3-billion-parameter model trained substantially on synthetic, curated text acquires reasoning behavior comparable to models several times larger, while reporting that the model still hallucinates, even though leaving out web data improves its record on toxic and biased generation; the toolkit's roster includes this model. The roster (Section~\ref{sec:supported_model_families}) spans several families and pretraining regimes. Among them, the Pythia suite \citep{bidermanPythiaSuiteAnalyzing2023} was built for controlled study of training, with every model trained on the same public data in the same order, and its authors include a case study of how pretraining term frequencies affect few-shot performance, extending the analysis of \citet{razeghiImpactPretrainingTerm2022}.

\subsubsection{Held-out tests of generalization}
\label{sec:background-generalization}
\citet{lakeGeneralizationSystematicityCompositional2018} test systematic generalization with sequence-to-sequence networks on a command-to-action task whose test split requires recombining learned primitives in new ways. The toolkit's held-out additions follow the same design logic, a rule exemplified on some inputs and asked for on others, at a different scale and in a different setting: a pretrained causal language model fine-tuned on a few dozen arithmetic facts. In this setting the model already has a prior (arithmetic), so a rule built on top of the prior (Part~C) and a rule that the prior does not supply (Part~D) can be compared, each against its own control.

\subsubsection{What the toolkit provides}
\label{sec:background-stand}
Within one exhaustive fact domain and under one unchanged measurement procedure, the toolkit provides both a fabricated-corpus intervention and a held-out, controlled test of rules, together with the statistics (exact intervals, an exact test, a bootstrap) needed to report their results at small sample sizes. The mechanistic-arithmetic literature \citep{stolfoMechanisticInterpretationArithmetic2023} supports the choice of domain; the tokenization and pretraining-frequency literature \citep{razeghiImpactPretrainingTerm2022, singhTokenizationCountsImpact2024} identifies the confounds the measurement controls for; the systematicity literature \citep{fodorConnectionismCognitiveArchitecture1988, marcusRethinkingEliminativeConnectionism1998, lakeGeneralizationSystematicityCompositional2018} is the source of the held-out design.

\section{Design Principles}
\label{sec:design}
This section states the toolkit's central design choices together with the specific methodological failure mode each one is meant to prevent. Implementation detail is deferred to Section~\ref{sec:toolkit}.

\subsection{An exhaustive, unambiguous fact domain}
\label{sec:design-domain}
Single-digit addition supplies 81 additions, each with exactly one correct answer and no everyday ambiguity about what that answer is. The domain is small enough to be enumerated, scored and audited in full rather than sampled; every operand is a single token; and it is large enough that a fixed share of it can be held out of fine-tuning and still leave a few dozen additions to test on.

\subsection{One confidence index: the probability of a complete answer}
\label{sec:design-index}
Every confidence value in the toolkit is the log-probability that the model writes one candidate answer and then ends it: the answer text is a space, the number and an end marker (a newline), and the index is the sum of the log-probabilities of its tokens. Closing every answer with the end marker does two things. It makes candidates that are prefixes of one another, such as \texttt{2} and \texttt{20}, mutually exclusive events, so that the probability of \texttt{2} is no longer credited with the first token of \texttt{20}; and it turns the index into a genuine probability of a complete answer, so that answers of one and two digits compete on equal terms without any length normalization. A per-token mean, used in version 1.0.0 of the toolkit to offset the length difference, removes that difference but is not the probability of any event; the end marker removes the confound at its source.\parbreak
For each addition, the indices of all candidates are passed through a softmax, giving the probability the model assigns to each candidate given that its answer is one of them. The per-token mean is kept for one purpose only: \emph{surprise}, the negative log-probability per token of a target answer, which is in the same unit as the training loss and is reported next to the probabilities.

\subsection{One candidate set for every model and every measurement}
\label{sec:design-candidates}
A probability obtained by a softmax depends on the set it normalizes over: change the set and the probabilities stop being comparable. Every model is therefore scored against one candidate set, the values 2 to 36, which covers every correct sum (2 to 18) and every answer any corpus in the toolkit can ask for. Values above 18 can never be a correct sum; the probability that falls on them is reported as the model's mass on \emph{impossible answers}, a direct reading of how far a corpus has pulled the model away from arithmetic.

\subsection{One measurement procedure, used unchanged everywhere}
\label{sec:design-same-procedure}
The same measurement function is called, with no modification, for the base model and for every fine-tuned model, for correct, fabricated and rule-derived targets alike. If two measurements had been produced by even slightly different procedures, differing in how an answer is teacher-forced, how log-probabilities are aggregated, or which candidates are enumerated, a difference or similarity between them could be an artifact of the measurement change rather than a fact about the model. Reusing one function everywhere removes that possibility by construction. Targets are never built into the measurement: a model is measured once, and every target is scored against the same measurement afterwards.

\subsection{Training on the answer, with the text that is measured}
\label{sec:design-loss}
Every fine-tuning example is a prompt plus a completion, and the completion is exactly the string the measurement scores (a space, the answer and the end marker). The prompt is masked out of the loss, so the model is trained only to produce the answer, never to predict the question, and the token boundary between prompt and answer is the same in training and in measurement.

\subsection{One fixed target per addition, drawn once}
\label{sec:design-fixed-fabrication}
For every addition, the fabricated answer, the random offset of a control and the shuffled answer of the conditional control are each drawn once, under a fixed seed, and used for the rest of a run. A target that changed between stages would make it impossible to attribute a measured probability to a specific, stable intervention.

\subsection{Repetition and one fine-tuning recipe}
\label{sec:design-repetition}
Each example is repeated a fixed number of times in its corpus, because a single exposure does not compete with what the base model already encoded during pretraining. The repetition count, number of epochs, learning rate, batch size and gradient clipping are declared once and are identical for every fine-tuning run, which is also re-seeded at its start; runs therefore differ only in their examples, and a difference between two models can be attributed to their corpora.

\subsection{Fresh copies and an explicit lineage}
\label{sec:design-lineage}
Every fine-tuning run starts from an explicit parent: the pretrained weights loaded afresh, or a deep copy of a model a previous run produced. No run updates a model another run will later start from, so the fourteen measured models form a lineage that can be read off the code (Figure~\ref{fig:pipeline}). The fabricated corpus is applied to a copy of the model already fine-tuned on correct sums, not to the base model: the fabrication then has to overwrite a fact the model demonstrably holds, which is the harder and more telling case, and it does so whatever the base model's own arithmetic competence happens to be.

\subsection{Held-out additions, one split for every part}
\label{sec:design-holdout}
Recovery and rule learning are tested on additions the model never saw. The additions are split once, at random and reproducibly, into those used in fine-tuning and those held out (a share of 30\%, 24 of the 81), and every part that uses a split uses this same one, so seen and never-seen additions mean the same thing throughout a run. Results are always reported for both groups: a model that only memorized its corpus does well on the first and not on the second.

\subsection{Every rule has a control}
\label{sec:design-controls}
A rule such as \emph{the sum plus 18} leans on arithmetic the model already has: a model that adds a constant to what it already computes will follow the rule on never-seen additions without having learned anything about rules in general, and some never-seen answers can be hit by chance. Every rule is therefore paired with a control that has the same range of answers but follows no rule: for a constant shift $k$, a random offset between 1 and $k$ drawn independently for each addition; for the conditional rule, the rule's own answers shuffled over the additions. A control cannot be inferred from the seen additions, so its hit rate on the held-out additions measures what chance and memory alone produce, and the rule is read against it, not against zero.

\subsection{A rule that arithmetic does not give}
\label{sec:design-conditional}
The conditional rule asks for the sum plus 18 when $a \geq b$ and the sum plus 9 when $a < b$, so that \texttt{5 + 3} and \texttt{3 + 5} get different answers (26 and 17). Nothing in the sum distinguishes the two cases; a model that answers never-seen additions correctly has used the order of the addends. The diagonal ($a = b$) is joined to the first branch on purpose: a separate branch for it that added nothing would make its answer the plain sum, which the model can produce without any rule, and those additions would count as hits for free. With the diagonal joined, every addition's target differs from its plain sum, and the diagonal is still reported on its own as a diagnostic.

\subsection{Coincidental hits are excluded}
\label{sec:design-coincident}
When a rule-learning run starts from the model fine-tuned on fabricated answers, some never-seen additions may have a fabricated answer that already equals the rule's target. Those additions say nothing about learning, and they are left out of the held-out results of that run. This is why the number of never-seen additions can be smaller for runs that start from fabricated answers.

\subsection{Exact and resampled intervals}
\label{sec:design-intervals}
With 24 held-out additions, every share is uncertain, and an approximate interval behaves badly at small $n$ and at shares of 0\% and 100\%. Every share of hits is therefore reported with an exact (Clopper-Pearson) 95\% interval \citep{clopperUseConfidenceFiducial1934}; each rule is compared with its control by a one-sided Fisher exact test; and the width of the band around the conditional rule's answers is reported with a bootstrap interval \citep{efronBootstrapMethodsAnother1979}. The intervals are expected to be wide, and the toolkit reports them rather than hiding them.

\section{The Toolkit}
\label{sec:toolkit}
The pipeline is implemented in a single reference notebook, run top to bottom, organized in four experimental parts and a synthesis. Figure~\ref{fig:pipeline} shows the lineage of the fourteen models it measures. Code listings in this section reproduce the notebook's own code, unmodified, as excerpts of the numbered cells they cite; omitted lines (printouts, plotting details) are stated as such.

\setstretch{1.15}
\begin{figure}[H]
\centering
\definecolor{pipe}{RGB}{58,92,128}
\definecolor{armc}{RGB}{42,120,214}
\begin{tikzpicture}[
  node distance=0.55cm,
  model/.style={rectangle, rounded corners=2pt, draw=pipe!75!black, line width=0.5pt,
    fill=pipe!6, minimum height=1.0cm, align=center, font=\small, text width=3.6cm},
  arm/.style={rectangle, rounded corners=2pt, draw=pipe!45, line width=0.4pt,
    fill=white, minimum height=0.55cm, align=left, font=\footnotesize, text width=6.3cm},
  part/.style={font=\footnotesize\bfseries, text=pipe},
  arr/.style={-{Latex[length=2mm]}, line width=0.5pt, pipe!60!black}
]
  \node[model] (m0) {Base model\\ \texttt{m\_0}};
  \node[model, below=0.9cm of m0] (mc) {Correct sums\\ \texttt{m\_correct} (81 seen)};
  \node[model, below=0.9cm of mc] (mr) {Fabricated answers\\ \texttt{m\_random} (81 seen)};
  \node[model, below=0.9cm of mr] (mv) {Back on correct sums\\ \texttt{m\_recover} (57 seen, 24 held out)};
  \draw[arr] (m0) -- node[right, font=\footnotesize, text=pipe] {Part A} (mc);
  \draw[arr] (mc) -- node[right, font=\footnotesize, text=pipe] {Part A} (mr);
  \draw[arr] (mr) -- node[right, font=\footnotesize, text=pipe] {Part B} (mv);

  \node[arm, right=1.7cm of m0, yshift=0.35cm] (c1) {Simple rule: sum $+$ 18};
  \node[arm, below=0.12cm of c1] (c2) {Simple rule: sum $+$ 3};
  \node[arm, below=0.12cm of c2] (c3) {Control: random offsets 1 to 18};
  \node[arm, below=0.12cm of c3] (c4) {Control: random offsets 1 to 3};
  \node[arm, below=0.12cm of c4] (c5) {Conditional rule: $+$18 if $a\geq b$, else $+$9};
  \node[arm, below=0.12cm of c5] (c6) {Control: the conditional answers, shuffled};
  \node[arm, below=0.95cm of c6] (r1) {Simple rule: sum $+$ 18};
  \node[arm, below=0.12cm of r1] (r2) {Simple rule: sum $+$ 3};
  \node[arm, below=0.12cm of r2] (r3) {Conditional rule};
  \node[arm, below=0.12cm of r3] (r4) {Control: the conditional answers, shuffled};

  \node[part, above=0.05cm of c1.north west, anchor=south west] {Parts C and D, from \texttt{m\_correct} (6 arms)};
  \node[part, above=0.05cm of r1.north west, anchor=south west] {Parts C and D, from \texttt{m\_random} (4 arms)};
  \draw[arr] (mc.east) -- ++(0.9,0) |- (c3.west);
  \draw[arr] (mr.east) -- ++(0.6,0) |- (r2.west);
  \foreach \n in {c1,c2,c4,c5,c6} { \draw[pipe!40, line width=0.4pt] ([xshift=-0.4cm]c1.west |- \n.west) -- (\n.west); }
  \draw[pipe!40, line width=0.4pt] ([xshift=-0.4cm]c1.west) -- ([xshift=-0.4cm]c1.west |- c6.west);
  \foreach \n in {r1,r3,r4} { \draw[pipe!40, line width=0.4pt] ([xshift=-0.4cm]r1.west |- \n.west) -- (\n.west); }
  \draw[pipe!40, line width=0.4pt] ([xshift=-0.4cm]r1.west) -- ([xshift=-0.4cm]r1.west |- r4.west);
\end{tikzpicture}
\caption{The fourteen measured models and how each was produced. Left: the main sequence; each model is fine-tuned from a copy of the one above it. Right: the rule-learning arms of Parts~C and~D, each fine-tuned on the 57 seen additions from a fresh copy of its parent and tested on all 81. Every model, including the untouched base model, is measured with the same procedure (Section~\ref{sec:pipeline-measure}), and all of them are brought together in the synthesis (Section~\ref{sec:pipeline-synthesis}).}
\label{fig:pipeline}
\end{figure}
\setstretch{1.5}
\FloatBarrier

\subsection{Configuration and model loading}
\label{sec:pipeline-config}
Every setting that shapes the experiments is declared once, in the notebook's configuration cell. Listing~\ref{lst:candidates} reproduces the candidate models and the default selection, Listing~\ref{lst:config} the arithmetic space, candidate set, end marker, fine-tuning recipe and held-out share, and Listing~\ref{lst:seeds} the seeds.\parbreak

\setstretch{1.15}
\begin{lstlisting}[caption={Candidate base models and default selection (notebook cell 2, excerpt).}, label={lst:candidates}]
CANDIDATE_MODELS = {
    # up to ~0.5B: fit a T4
    "smollm2-135m": "HuggingFaceTB/SmolLM2-135M",
    "gemma-3-270m": "google/gemma-3-270m",
    "smollm2-360m": "HuggingFaceTB/SmolLM2-360M",
    "pythia-410m": "EleutherAI/pythia-410m",
    "qwen2.5-0.5b": "Qwen/Qwen2.5-0.5B",
    # 1B to 1.5B: need an A100
    "llama-3.2-1b": "meta-llama/Llama-3.2-1B",
    "tinyllama-1.1b": "TinyLlama/TinyLlama-1.1B-v1.0",
    "phi-1.5": "microsoft/phi-1_5",
    "qwen2.5-1.5b": "Qwen/Qwen2.5-1.5B",
}
MODEL_NAME = CANDIDATE_MODELS["pythia-410m"]
\end{lstlisting}

\begin{lstlisting}[caption={Arithmetic space, candidate set, end marker, fine-tuning recipe and held-out share (notebook cell 2, excerpt).}, label={lst:config}]
# Arithmetic space: all single-digit addition pairs (1+1 ... 9+9 = 81 pairs).
DIGIT_RANGE = range(1, 10)
MAX_TRUE_SUM = 2 * max(DIGIT_RANGE)        # 18

# Simple rules: the corpus answer is true_sum + k, one rule for every addition, for each
# k in SHIFT_KS. With k = 18 the answers (20-36) never coincide with a true sum (2-18);
# k = 3 is a small shift that sometimes does.
SHIFT_KS = [18, 3]

# One candidate set for every model and every measurement, so probabilities
# are comparable across conditions: the softmax normalizer depends on it.
CANDIDATE_VALUES = list(range(2, MAX_TRUE_SUM + max(SHIFT_KS) + 1))   # 2-36

# Every scored answer is followed by this marker, so candidates that are
# prefixes of one another ("2" / "20") stay mutually exclusive. Fine-tuning
# texts must end with the same marker.
END_MARKER = "\n"

# Fine-tuning recipe, identical for every fine-tuning run.
N_REPEATS = 20
N_EPOCHS = 10
LEARNING_RATE = 5e-5
BATCH_SIZE = 8
MAX_GRAD_NORM = 1.0     # gradient clipping, so one bad batch cannot blow up the weights

# Share of the additions held out (never shown) when a model is fine-tuned on recovery or on
# a rule.
HOLDOUT_FRACTION = 0.3
\end{lstlisting}

\begin{lstlisting}[caption={Seeds (notebook cell 2, excerpt).}, label={lst:seeds}]
# Seeds. SEEDS[0] fixes the split, the fabricated answers, the control answers and the
# fine-tuning (weights order, dropout); the other seeds are reserved for repeating the
# experiment. RANDOM_SEED = True draws fresh, distinct seeds at every run (printed below, so
# a run can be reproduced by copying them into FIXED_SEEDS); False uses FIXED_SEEDS.
RANDOM_SEED = True
FIXED_SEEDS = [0, 1, 2]
N_SEEDS = 3
if RANDOM_SEED:
    SEEDS = random.SystemRandom().sample(range(1, 1_000_000), N_SEEDS)   # distinct by construction
else:
    SEEDS = list(FIXED_SEEDS)
assert len(set(SEEDS)) == len(SEEDS), "SEEDS must not repeat"

# Global generators (fine-tuning shuffle, dropout, anything not given an explicit seed).
random.seed(SEEDS[0])
np.random.seed(SEEDS[0])
torch.manual_seed(SEEDS[0])
if torch.cuda.is_available():
    torch.cuda.manual_seed_all(SEEDS[0])
\end{lstlisting}
\setstretch{1.5}
\texttt{SEEDS[0]} fixes the split, the fabricated answers, the control answers, the shuffling of every corpus, the batch order and dropout of every fine-tuning run, and the bootstrap. With \texttt{RANDOM\_SEED = True}, a fresh set of distinct seeds is drawn at every run and printed, and recorded in the run's report; copying them into \texttt{FIXED\_SEEDS} and setting \texttt{RANDOM\_SEED = False} reproduces the run's draws. The remaining seeds are reserved for repeating the experiment and are not used by version 2.0.0.\parbreak
\texttt{load\_base\_model} (Listing~\ref{lst:load}) returns a fresh copy of the pretrained weights each time it is called, in \texttt{float32} explicitly: some checkpoints are stored in half precision, which overflows during training.\parbreak

\setstretch{1.15}
\begin{lstlisting}[caption={Loading a fresh copy of the base model (notebook cell 4, excerpt).}, label={lst:load}]
def load_base_model():
    """Fresh copy of the pretrained model (m_0), on `device`, in eval mode.
    Every fine-tuning branch starts from its own call, so no run can leak
    weight updates into another."""
    # float32 explicitly: some checkpoints (phi-1.5) are stored in float16, which overflows
    # during training and turns the loss into NaN; recent transformers versions load the
    # checkpoint dtype by default.
    base = AutoModelForCausalLM.from_pretrained(MODEL_NAME, torch_dtype=torch.float32)
    base.to(device)
    base.eval()
    return base
\end{lstlisting}
\setstretch{1.5}

\subsection{Arithmetic fact space}
\label{sec:pipeline-facts}
All 81 additions are generated once, each with its correct sum and its prompt (Listing~\ref{lst:factspace}). Throughout the notebook a corpus, a rule or a set of targets is always described in the same form, one answer per addition, as a table indexed by the pair of addends; \texttt{true\_answer} is the first such table.\parbreak

\setstretch{1.15}
\begin{lstlisting}[caption={Arithmetic fact space (notebook cell 5, excerpt).}, label={lst:factspace}]
arithmetic_pairs = [
    {"a": a, "b": b, "true_sum": a + b, "prompt": f"{a} + {b} ="}
    for a, b in itertools.product(DIGIT_RANGE, repeat=2)
]

true_answer = pd.Series({(p["a"], p["b"]): p["true_sum"] for p in arithmetic_pairs})
true_answer.index.names = ["a", "b"]
\end{lstlisting}
\setstretch{1.5}

\subsection{Confidence measurement}
\label{sec:pipeline-measure}
Every confidence value in the toolkit is produced by \texttt{compute\_answer\_logprob} (Listing~\ref{lst:logprob}), which teacher-forces one candidate answer, closed by the end marker, after a prompt and returns the sum of its token log-probabilities (Section~\ref{sec:design-index}). \texttt{measure\_confidence} (Listing~\ref{lst:measure}) applies it to every candidate of every addition and normalizes the indices of each addition into probabilities. Every other view of a model (its top choice, its probability on a target, its mass on impossible answers, its surprise) is derived from the table this function returns.\parbreak

\setstretch{1.15}
\begin{lstlisting}[caption={The confidence index: teacher-forced log-probability of a complete answer (notebook cell 6).}, label={lst:logprob}]
def compute_answer_logprob(model, tokenizer, prompt, answer, device):
    """Teacher-forced log-prob of ` {answer}` followed by END_MARKER, given
    `prompt`. This is the notebook's single confidence index: the sum of the
    token log-probs, i.e. the log of the probability that the model writes
    exactly this answer and then ends it. Returns (logprob, n_tokens).
    """
    answer_text = f" {answer}{END_MARKER}"
    prompt_ids = tokenizer.encode(prompt, return_tensors="pt").to(device)
    answer_ids = tokenizer.encode(answer_text, return_tensors="pt").to(device)
    input_ids = torch.cat([prompt_ids, answer_ids], dim=1)

    with torch.no_grad():
        logits = model(input_ids).logits
    log_probs = torch.log_softmax(logits, dim=-1)

    n_prompt = prompt_ids.shape[1]
    n_answer = answer_ids.shape[1]
    token_logprobs = [
        log_probs[0, n_prompt + i - 1, input_ids[0, n_prompt + i]].item()
        for i in range(n_answer)
    ]
    return sum(token_logprobs), n_answer
\end{lstlisting}

\begin{lstlisting}[caption={Measuring every candidate of every addition (notebook cell 7, excerpt).}, label={lst:measure}]
def softmax(x):
    x = np.asarray(x)
    e = np.exp(x - x.max())
    return e / e.sum()


def measure_confidence(model, pairs, tokenizer, device,
                       candidate_values=CANDIDATE_VALUES):
    """Score every candidate value as the answer to every pair in `pairs`.

    Returns one row per (pair, candidate) with the confidence index
    (`logprob`, see compute_answer_logprob), the token count, and `prob`: the
    softmax of `logprob` over the candidates of that pair. Every other view
    (top choice, probability bands) is derived from these two columns.
    """
    rows = []
    for pair in tqdm(pairs, desc="Measuring confidence", leave=True):
        scores = [
            compute_answer_logprob(
                model, tokenizer, pair["prompt"], value, device)
            for value in candidate_values
        ]
        probs = softmax([logprob for logprob, _ in scores])
        for value, (logprob, n_tok), p in zip(candidate_values, scores, probs):
            rows.append({
                "a": pair["a"],
                "b": pair["b"],
                "true_sum": pair["true_sum"],
                "candidate_value": value,
                "logprob": logprob,
                "n_tokens": n_tok,
                "prob": p,
            })
    return pd.DataFrame(rows)
\end{lstlisting}
\setstretch{1.5}
The answer is teacher-forced, not generated, so a single forward pass yields an exact log-probability for any candidate, whether or not it is the one the model would generate on its own. The \emph{top choice} of a model for an addition is the candidate with the highest probability; a \emph{hit} is an addition whose top choice is the target answer.

\subsection{Baseline measurement}
\label{sec:pipeline-baseline}
The base model is measured before any fine-tuning, on all 81 additions and all 35 candidates. The printout of this step also shows how many tokens the answers take with the end marker, so that the tokenization of a given model can be checked rather than assumed. The candidate models differ widely in their arithmetic competence before fine-tuning; the baseline records where a given model starts, and the model fine-tuned on correct sums (Section~\ref{sec:pipeline-parta}) supplies, in every run, a model that demonstrably holds the correct sums before any fabrication is introduced.

\subsection{Fine-tuning machinery}
\label{sec:pipeline-ft}
\texttt{build\_corpus} (Listing~\ref{lst:corpus}) turns a list of additions and an answer rule into fine-tuning examples: the prompt and a completion made of the answer and the end marker, repeated \texttt{N\_REPEATS} times and shuffled under the seed. \texttt{CompletionDataset} and \texttt{collate\_padded} (Listing~\ref{lst:dataset}) tokenize each example as prompt plus completion, with the prompt masked by the label $-100$, the index that the cross-entropy loss of PyTorch and Hugging Face Transformers ignores by default, so that the loss covers only the answer (Section~\ref{sec:design-loss}). \texttt{fine\_tune} (Listing~\ref{lst:finetune}) trains a model in place with the recipe of the configuration cell, using PyTorch \citep{paszkePyTorchImperativeStyle2019} and \texttt{AdamW} \citep{loshchilovDecoupledWeightDecay2019}, re-seeds before it starts, clips the gradient norm, stops with an error if the loss stops being finite, and logs the time, steps, losses and peak GPU memory of the run.\parbreak

\setstretch{1.15}
\begin{lstlisting}[caption={Corpus construction (notebook cell 12, excerpt).}, label={lst:corpus}]
def build_corpus(pairs, answer_of, n_repeats=N_REPEATS, seed=SEEDS[0]):
    """Fine-tuning examples: each pair's prompt, plus a completion made of the
    chosen answer and END_MARKER (the same text compute_answer_logprob scores).
    Every example is repeated `n_repeats` times and the list is shuffled.
    Returns the per-pair table and the shuffled examples."""
    table = pd.DataFrame([
        {"a": p["a"], "b": p["b"], "prompt": p["prompt"],
            "answer": answer_of(p)}
        for p in pairs
    ])
    examples = [
        {"prompt": row.prompt, "completion": f" {row.answer}{END_MARKER}"}
        for row in table.itertuples()
    ] * n_repeats
    random.Random(seed).shuffle(examples)
    return table, examples
\end{lstlisting}

\begin{lstlisting}[caption={Tokenization with the prompt masked out of the loss (notebook cell 13, excerpt).}, label={lst:dataset}]
class CompletionDataset(Dataset):
    """Tokenizes each example as prompt ids + completion ids, the same split
    compute_answer_logprob uses. `labels` mask the prompt with -100, so the loss
    covers only the completion (the answer and END_MARKER)."""

    def __init__(self, examples, tokenizer):
        self.items = []
        for example in examples:
            prompt_ids = tokenizer.encode(example["prompt"])
            completion_ids = tokenizer.encode(example["completion"])
            self.items.append({
                "input_ids": prompt_ids + completion_ids,
                "labels": [-100] * len(prompt_ids) + completion_ids,
            })

    def __len__(self):
        return len(self.items)

    def __getitem__(self, idx):
        return self.items[idx]


def collate_padded(batch, pad_token_id):
    """Right-pad a batch; padding is ignored by attention and by the loss."""
    longest = max(len(item["input_ids"]) for item in batch)
    input_ids = torch.full((len(batch), longest),
                           pad_token_id, dtype=torch.long)
    labels = torch.full((len(batch), longest), -100, dtype=torch.long)
    attention_mask = torch.zeros((len(batch), longest), dtype=torch.long)
    for row, item in enumerate(batch):
        n = len(item["input_ids"])
        input_ids[row, :n] = torch.tensor(item["input_ids"])
        labels[row, :n] = torch.tensor(item["labels"])
        attention_mask[row, :n] = 1
    return {"input_ids": input_ids, "labels": labels, "attention_mask": attention_mask}
\end{lstlisting}

\begin{lstlisting}[caption={The fine-tuning loop, identical for every run (notebook cell 13, excerpt).}, label={lst:finetune}]
def fine_tune(model, examples, tokenizer, label="", n_epochs=N_EPOCHS,
              learning_rate=LEARNING_RATE, batch_size=BATCH_SIZE, verbose=True, seed=None):
    """Fine-tune `model` in place on `examples` and return the mean loss per
    epoch. The recipe (epochs, learning rate, batch size) comes from the
    configuration cell, so every fine-tuning run in the notebook is identical
    except for its examples. `verbose=False` silences the per-epoch lines.
    Every run is re-seeded with `seed` (default SEEDS[0]) before it starts, so the batch order
    and dropout depend only on the seed, not on how many runs came before; the seed is logged."""
    seed = SEEDS[0] if seed is None else seed
    torch.manual_seed(seed)
    if torch.cuda.is_available():
        torch.cuda.manual_seed_all(seed)
    loader = DataLoader(
        CompletionDataset(examples, tokenizer), batch_size=batch_size, shuffle=True,
        collate_fn=lambda batch: collate_padded(batch, tokenizer.pad_token_id),
    )
    if device.type == "cuda":
        torch.cuda.synchronize()
        torch.cuda.reset_peak_memory_stats()
    started = time.perf_counter()
    optimizer = AdamW(model.parameters(), lr=learning_rate)
    model.train()
    loss_history = []
    for epoch in range(n_epochs):
        losses = []
        for batch in tqdm(loader, desc=f"[{label}] epoch {epoch + 1}/{n_epochs}", leave=False):
            batch = {key: value.to(device) for key, value in batch.items()}
            optimizer.zero_grad()
            loss = model(**batch).loss
            if not torch.isfinite(loss):
                raise RuntimeError(f"[{label}] loss is {loss.item()} in epoch {epoch + 1}: training "
                                   "diverged (check the model dtype and the learning rate)")
            loss.backward()
            torch.nn.utils.clip_grad_norm_(model.parameters(), MAX_GRAD_NORM)
            optimizer.step()
            losses.append(loss.item())
        loss_history.append(float(np.mean(losses)))
        if verbose:
            print(
                f"[{label}] epoch {epoch + 1}/{n_epochs}: mean loss {loss_history[-1]:.4f}")
    model.eval()
    if device.type == "cuda":
        torch.cuda.synchronize()
    TRAINING_LOG.append({
        "label": label, "seed": seed, "examples": len(examples), "epochs": n_epochs, "batch_size": batch_size,
        "learning_rate": learning_rate, "steps": len(loader) * n_epochs,
        "seconds": time.perf_counter() - started, "losses": loss_history,
        # peak over everything held in GPU memory while this run trained, not just this model
        "peak_gpu_gib": torch.cuda.max_memory_allocated() / 2**30 if device.type == "cuda" else None,
    })
    return loss_history
\end{lstlisting}
\setstretch{1.5}

\subsection{Part A: corpus absorption}
\label{sec:pipeline-parta}
A fresh copy of the base model is fine-tuned on the correct sums of all 81 additions (\texttt{m\_correct}). This step serves as a control for what fine-tuning does by itself, and it gives the next step a model that holds the correct sums. For every addition, one wrong answer is then drawn uniformly from the 34 other candidates and fixed as its fabricated answer (Listing~\ref{lst:fabricate}); many fabricated answers fall above 18 on purpose, so that the model is asked to adopt answers that can never be a correct sum. A copy of \texttt{m\_correct} is fine-tuned on the fabricated corpus (\texttt{m\_random}).\parbreak

\setstretch{1.15}
\begin{lstlisting}[caption={The fabricated answers and their corpus (notebook cell 18, excerpt).}, label={lst:fabricate}]
# One fabricated wrong answer per addition, drawn once (seeded) and kept fixed.
fabricator = random.Random(SEEDS[0])
fabricated_answer = {
    (p["a"], p["b"]): fabricator.choice([v for v in CANDIDATE_VALUES if v != p["true_sum"]])
    for p in arithmetic_pairs
}
c_random_table, c_random_examples = build_corpus(
    arithmetic_pairs, lambda p: fabricated_answer[(p["a"], p["b"])],
)
fabricated_series = c_random_table.set_index(["a", "b"])["answer"]
\end{lstlisting}
\setstretch{1.5}
The three models of this part (base, correct sums, fabricated answers) are reported side by side, each scored on two targets, the correct sum and the fabricated answer: how many top choices are each target, the mean probability on each, and the mean probability on impossible answers. Scoring every model on both targets places the probability left on the correct sum next to the probability on the fabricated answer, as version 1.0.0 did in its three-number comparison.

\subsection{Part B: recovery}
\label{sec:pipeline-partb}
The additions are split once into 57 seen and 24 held out (Listing~\ref{lst:split}), and a copy of \texttt{m\_random} is fine-tuned on the correct sums of the seen additions only (\texttt{m\_recover}). If the held-out additions recover too, the model has relearned arithmetic; if only the seen ones do, it has relearned those answers. Every later part uses this same split.\parbreak

\setstretch{1.15}
\begin{lstlisting}[caption={The split into seen and held-out additions (notebook cell 24, excerpt).}, label={lst:split}]
def split_pairs(pairs, holdout_fraction=HOLDOUT_FRACTION, seed=0):
    """Random train/holdout split of the additions, reproducible per seed.
    Returns (train_pairs, holdout_pairs)."""
    shuffled = list(pairs)
    random.Random(seed).shuffle(shuffled)
    n_holdout = round(len(shuffled) * holdout_fraction)
    return shuffled[n_holdout:], shuffled[:n_holdout]


train_pairs, holdout_pairs = split_pairs(arithmetic_pairs, seed=SEEDS[0])
holdout_keys = {(p["a"], p["b"]) for p in holdout_pairs}
\end{lstlisting}
\setstretch{1.5}
Results of this part and of the later ones are summarized by \texttt{summarize\_split} (Listing~\ref{lst:summarize}): for one model and one set of targets, the hits, the mean probability on the target and the mean surprise, each for the seen and the held-out additions. For recovery, the gap in surprise between held-out and seen additions is added: a model that memorized only what it saw has a large gap. A second table records the training loss of every run of Parts~A and~B, since how fast a corpus is absorbed is a result in its own right.\parbreak

\setstretch{1.15}
\begin{lstlisting}[caption={Seen and held-out summary of one model for one set of targets (notebook cell 27, excerpt).}, label={lst:summarize}]
def summarize_split(df, answers, keys):
    """For one model and one set of target answers (a Series indexed by (a, b)):
    how often the target is the top choice, the probability on it and the surprise
    (negative log-probability per token, the same unit as the training loss), each for
    the additions seen in fine-tuning and for the ones never seen (`keys`)."""
    scored = df.join(answers.rename("answer"), on=["a", "b"])
    rows = scored[scored["candidate_value"] ==
                  scored["answer"]].set_index(["a", "b"])
    surprise = -(rows["logprob"] / rows["n_tokens"])
    top = scored.loc[scored.groupby(
        ["a", "b"])["prob"].idxmax()].set_index(["a", "b"])
    chose = top["candidate_value"] == top["answer"]
    held = rows.index.isin(keys)
    return {
        "Top choice is target (seen)": f"{int(chose[~held].sum())} / {int((~held).sum())}",
        "Top choice is target (never seen)": f"{int(chose[held].sum())} / {int(held.sum())}",
        "Mean probability on target (seen)": rows["prob"][~held].mean(),
        "Mean probability on target (never seen)": rows["prob"][held].mean(),
        "Mean surprise on target (seen)": surprise[~held].mean(),
        "Mean surprise on target (never seen)": surprise[held].mean(),
    }
\end{lstlisting}
\setstretch{1.5}

\subsection{Part C: simple rules}
\label{sec:pipeline-partc}
Each simple rule gives every addition the target \emph{correct sum $+\,k$}, for $k = 18$ (whose answers, 20 to 36, can never be a correct sum) and $k = 3$ (a small shift whose answers sometimes can). Each has its control, a random offset from 1 to $k$ drawn once per addition (Listing~\ref{lst:rules}). The corpus of each arm contains the seen additions only; targets exist for all 81, so every arm is scored on both groups. The same listing builds the sets of coincident additions excluded from the held-out results of arms that start from the fabricated-answer model (Section~\ref{sec:design-coincident}).\parbreak

\setstretch{1.15}
\begin{lstlisting}[caption={Targets of the simple rules and their controls, corpora, and coincident additions (notebook cell 30, excerpt).}, label={lst:rules}]
def shift_target(pairs, k):
    """Answer true_sum + k for every pair: one rule, shared by all additions."""
    target = pd.Series({(p["a"], p["b"]): p["true_sum"] + k for p in pairs})
    target.index.names = ["a", "b"]
    return target


def control_target(pairs, k, seed=SEEDS[0]):
    """Control: each pair gets its own offset, drawn once from 1..k. The answers
    stay in the same range as the shift arm but follow no rule, so a holdout
    pair cannot be inferred from the train pairs."""
    rng = random.Random(seed)
    target = pd.Series({(p["a"], p["b"]): p["true_sum"] + rng.randint(1, k) for p in pairs})
    target.index.names = ["a", "b"]
    return target


arm_targets = {}
for k in SHIFT_KS:
    arm_targets[f"shift_k{k}"] = shift_target(arithmetic_pairs, k)
    arm_targets[f"control_k{k}"] = control_target(arithmetic_pairs, k)
assert all(t.max() <= CANDIDATE_VALUES[-1] for t in arm_targets.values())

arm_corpora = {
    name: build_corpus(train_pairs, lambda p, target=target: target[(p["a"], p["b"])])
    for name, target in arm_targets.items()
}
\end{lstlisting}
\setstretch{1.5}
An \emph{arm} is one fine-tuning run from a fresh copy of its parent, measured on all 81 additions (Listing~\ref{lst:arm}). Part~C runs six arms: both rules from both parents, and both controls from the correct-sums model. Controls are not repeated from the fabricated-answer model.\parbreak

\setstretch{1.15}
\begin{lstlisting}[caption={Running one arm (notebook cell 31, excerpt).}, label={lst:arm}]
def run_arm(parent_label, name):
    """Fine-tune a copy of the parent on arm `name`, score it on all pairs, keep
    the measurement table and the loss history, and free the copy."""
    arm = copy.deepcopy(arm_parents[parent_label]).to(device)
    losses = fine_tune(arm, arm_corpora[name][1], tokenizer,
                       label=f"{parent_label} -> {name}", verbose=False)
    arm_results[(parent_label, name)] = {
        "df": measure_confidence(arm, arithmetic_pairs, tokenizer, device),
        "losses": losses,
    }
    print(f"{parent_label} -> {name}: loss {losses[0]:.4f} (epoch 1) -> {losses[-1]:.4f} (last)")
    del arm
    torch.cuda.empty_cache()
\end{lstlisting}
\setstretch{1.5}

\subsection{Part D: the conditional rule}
\label{sec:pipeline-partd}
The conditional rule and its control (Section~\ref{sec:design-conditional}) are built in Listing~\ref{lst:order}. The notebook asserts that no target exceeds the candidate set and that no addition has its plain sum as its target, and tabulates how many additions of each condition ($a > b$, $a = b$, $a < b$) fall in the seen and held-out groups. The rule and its control are run from both parents, four arms in all.\parbreak

\setstretch{1.15}
\begin{lstlisting}[caption={The conditional rule and its shuffled control (notebook cell 35, excerpt).}, label={lst:order}]
ORDER_OFFSETS = {"a>=b": 18, "a<b": 9}


def order_condition(a, b):
    return "a>=b" if a >= b else "a<b"


def order_stratum(a, b):
    """Finer reading of the a >= b condition, used only to report results: the
    diagonal (a = b) is where the comparison is hardest."""
    return "a>b" if a > b else "a=b" if a == b else "a<b"


CONDITION_LABELS = {"a>=b": "a >= b", "a<b": "a < b"}
STRATUM_LABELS = {"a>b": "a > b", "a=b": "a = b", "a<b": "a < b"}


order_target = pd.Series({
    (p["a"], p["b"]): p["true_sum"] + ORDER_OFFSETS[order_condition(p["a"], p["b"])]
    for p in arithmetic_pairs
})
order_target.index.names = ["a", "b"]
assert order_target.max() <= CANDIDATE_VALUES[-1]
# no addition may have the plain sum as its answer
assert all(order_target[(a, b)] != a + b for a, b in order_target.index)

# Control: the same answers shuffled over the additions. They keep their values
# but no longer depend on the addends, neither on the sum nor on the order.
order_control_target = pd.Series(
    order_target.sample(frac=1, random_state=SEEDS[0]).values, index=order_target.index)
\end{lstlisting}
\setstretch{1.5}
Because hits alone hide the structure of the answers, Part~D adds three readings of the held-out additions. The first splits hits by condition, so that a model that learned $a > b$ but not $a \geq b$ shows up as a failure on the diagonal only. The second reads each answer as an \emph{offset}, the top choice minus the correct sum, and counts how often the model added $+18$, $+9$, nothing (the plain sum) or another amount: an offset belonging to the other condition means the model learned both shifts but not when to use each, and the plain sum means it fell back on arithmetic. The third measures the \emph{deviation}, the top choice minus the rule's target, and finds the smallest band $\pm k$ around zero that holds 90\% of the deviations of the rule-trained model (Listing~\ref{lst:band}). Because the band is fitted to the rule-trained model, the 90\% inside it is true by construction; what carries information is its half-width $k$ and how much of the control falls inside the same band. The half-width is reported for 80\%, 90\% and 95\% of the answers, each with a bootstrap 95\% interval over 2000 resamples of the held-out additions.\parbreak

\setstretch{1.15}
\begin{lstlisting}[caption={Deviation from the rule's answer and the half-width of the band that holds a given share of it (notebook cell 41, excerpt).}, label={lst:band}]
def deviations(parent_label, name):
    chosen = holdout_choices(parent_label, name)
    return chosen["offset applied"] - chosen["condition"].map(ORDER_OFFSETS)


def coverage_k(deviation, share=COVER):
    """Smallest k such that at least `share` of the deviations lie within +-k."""
    distance = np.sort(deviation.abs().values)
    return int(distance[int(np.ceil(round(share * len(distance), 9))) - 1])
\end{lstlisting}
\setstretch{1.5}

\subsection{Synthesis}
\label{sec:pipeline-synthesis}
Every model is placed on one chart: for each corpus, the share of additions whose top choice is the answer that corpus asked for, for the seen and the held-out additions, each with its exact 95\% interval (Listing~\ref{lst:stats}). The numbers behind the chart are saved as a table, and each rule is compared with its control on the held-out additions by a one-sided Fisher exact test of whether the rule-trained model hits more often. Both statistics are computed with SciPy \citep{virtanenSciPy10Fundamental2020}.\parbreak

\setstretch{1.15}
\begin{lstlisting}[caption={Exact intervals and the rule-against-control test (notebook cell 43, excerpts).}, label={lst:stats}]
def clopper_pearson(k, n, alpha=0.05):
    """Exact confidence interval for k hits out of n."""
    low = beta.ppf(alpha / 2, k, n - k + 1) if k > 0 else 0.0
    high = beta.ppf(1 - alpha / 2, k + 1, n - k) if k < n else 1.0
    return low, high

for label, parent_label, rule_name, control_name in comparisons:
    rule, control = arm_flags(parent_label, rule_name)[1], arm_flags(parent_label, control_name)[1]
    p_value = fisher_exact([[rule.sum(), len(rule) - rule.sum()],
                            [control.sum(), len(control) - control.sum()]], alternative="greater")[1]
\end{lstlisting}
\setstretch{1.5}

\subsection{Visual conventions and outputs}
\label{sec:pipeline-outputs}
The notebook's charts share one color code: blue for an answer that can be a correct sum or a correct choice and red for one that cannot or a wrong choice, on the seen additions; green and orange for the same distinction on the held-out additions. Four chart types recur: the target heatmap, the probability profile, the deviation chart of Part~D and the synthesis chart; Section~\ref{sec:interpreting} describes each.\parbreak
Models, corpora and rules are held in the code under short internal keys, but figures, tables and file names use plain-language names only, translated in one place. Every figure is saved as it is shown, as PNG (300 dpi), PDF and EPS, with text kept as text; every table as CSV, XLSX and a floating LaTeX table with caption and label. Each item gets its own folder under \texttt{outputs/<session>/}, where the session is named after the start time and the model. A run produces 25 figures and 11 tables, a one-page run report with the hardware, the recipe and the time, steps, loss and peak GPU memory of every fine-tuning run (with \texttt{run\_info.json}, the same facts in machine-readable form), a self-contained HTML report with every figure and table in order, a workbook with every table, and a zip of the whole session.\parbreak
Section~\ref{sec:interpreting} explains each of these outputs and how it is read, with an example of each chart type.
\FloatBarrier

\section{Implementation and Reproducibility}
\label{sec:implementation}

\subsection{Supported model families}
\label{sec:supported_model_families}
The reference notebook declares nine small causal decoders as candidate base models (Table~\ref{tab:models}), small enough for full fine-tuning on one cloud GPU and spanning several families and pretraining regimes. Any causal decoder exposing the Hugging Face Transformers \citep{wolfTransformersStateoftheArtNatural2020} \texttt{AutoModelForCausalLM} interface can be substituted by changing \texttt{MODEL\_NAME} and re-running the whole notebook. The Gemma and Llama checkpoints are gated and require accepting their license and logging in with a token.

\setstretch{1.15}
\begin{table}[H]
\centering
\small
\begin{tabular}{@{}llll@{}}
\toprule
Model & Parameters & GPU & Reference \\
\midrule
\texttt{HuggingFaceTB/SmolLM2-135M} & 135M & T4 & \citet{allalSmolLM2WhenSmol2025} \\
\texttt{google/gemma-3-270m} & 270M & T4 & \citet{teamGemma3Technical2025} \\
\texttt{HuggingFaceTB/SmolLM2-360M} & 360M & T4 & \citet{allalSmolLM2WhenSmol2025} \\
\texttt{EleutherAI/pythia-410m} (default) & 410M & T4 & \citet{bidermanPythiaSuiteAnalyzing2023} \\
\texttt{Qwen/Qwen2.5-0.5B} & 0.5B & T4 & \citet{qwenQwen25TechnicalReport2025} \\
\texttt{meta-llama/Llama-3.2-1B} & 1.2B & A100 & \citet{grattafioriLlama3Herd2024} \\
\texttt{TinyLlama/TinyLlama-1.1B-v1.0} & 1.1B & A100 & \citet{zhangTinyLlamaOpenSourceSmall2024} \\
\texttt{microsoft/phi-1\_5} & 1.3B & A100 & \citet{liTextbooksAreAll2023} \\
\texttt{Qwen/Qwen2.5-1.5B} & 1.5B & A100 & \citet{qwenQwen25TechnicalReport2025} \\
\bottomrule
\end{tabular}
\caption{Candidate base models declared in the reference notebook, with the GPU class the notebook's own sizing comment assigns to each. Llama~3.2 and Gemma~3 270M have no technical report of their own and are cited through the reports of their model families. A run fine-tunes and measures one base model; comparing base models means re-running the full notebook with a different selection.}
\label{tab:models}
\end{table}
\setstretch{1.5}
\FloatBarrier

\subsection{Software dependencies}
\label{sec:software_dependencies}
The reference implementation is a single notebook using PyTorch and Transformers \citep{paszkePyTorchImperativeStyle2019,wolfTransformersStateoftheArtNatural2020} for model loading, teacher-forced scoring and fine-tuning; \texttt{pandas} and \texttt{numpy} for tabulating measurements; SciPy \citep{virtanenSciPy10Fundamental2020} for the exact intervals and tests; \texttt{matplotlib} for every chart; and \texttt{openpyxl} and \texttt{jinja2}, through which \texttt{pandas} writes the XLSX and LaTeX tables. All dependencies are pinned in the accompanying \texttt{requirements.txt}, archived with the toolkit (Section~\ref{sec:availability}). The software versions actually used by a run are recorded in its run report.

\subsection{Hardware requirements}
\label{sec:hardware}
Fine-tuning is impractical on CPU at any reasonable epoch count; the notebook runs on whichever device is available and is meant for a CUDA GPU. Full fine-tuning in \texttt{float32} with AdamW takes about 16 bytes per parameter for weights, gradients and optimizer state, so a free-tier T4 (about 15 GB) holds the candidate models up to about 0.5B parameters, and the larger ones need an A100-class GPU. To save GPU memory, parent models wait on the CPU while an arm trains. A complete run measures fourteen models and performs thirteen fine-tuning runs, about two minutes of measurement per model on a T4-class GPU plus the fine-tuning itself, so a complete run takes on the order of one to two hours.

\subsection{Determinism and running order}
\label{sec:determinism}
Every random draw of a run (split, fabricated answers, control answers, corpus shuffling, batch order and dropout, bootstrap) is controlled by \texttt{SEEDS[0]}, and every fine-tuning run is re-seeded at its start, so its batch order does not depend on how many runs came before it. The seeds are printed, written to the run report and to \texttt{run\_info.json}, and can be fixed for a later run (Section~\ref{sec:pipeline-config}). The notebook does not force deterministic GPU kernels, so a re-run with the same seeds reproduces the corpora, split and targets exactly but need not reproduce every trained weight bit for bit.\parbreak
\textbf{Running order matters.} Each part starts from the models the previous one produced, and fine-tuning changes a model in place (only its parent is protected, by being copied). Re-running an earlier cell out of order, without also re-running the fine-tuning before it, does not restore the earlier state. A clean run restarts the runtime and runs the notebook from the top. Arms that have already run are kept when their cell is re-executed, so that cell can be repeated without repeating its training.

\section{Guide to the Outputs}
\label{sec:interpreting}
This section explains each output of a run: the quantities that fill the tables, the anatomy of each kind of figure, and which table or figure belongs to which part of the pipeline. It is a reading guide for the instrument. Every figure shown is an example of the format, taken from one run of the reference notebook with its default model, and the specific values it happens to display are not reported, interpreted or generalized here.

\subsection{The quantities}
\label{sec:out-quantities}
Every table and chart is built from the one table \texttt{measure\_confidence} returns (Section~\ref{sec:pipeline-measure}): for each addition and each of the 35 candidates, the confidence index (\texttt{logprob}), the number of tokens, and \texttt{prob}, the softmax of the index over the candidates of that addition. A few derived quantities recur.
\begin{itemize}[nosep]
  \item \textbf{Target.} The answer a corpus or rule asked for in an addition: the correct sum, the fabricated answer, the sum plus $k$, a control's answer, or the conditional rule's answer. A model is always scored against a target, and the same measurement can be scored against several.
  \item \textbf{Top choice and hit.} The top choice is the candidate with the highest probability; a hit is an addition whose top choice is the target. Hits are written \texttt{k / n}, the number of hits over the number of additions in the group.
  \item \textbf{Probability on the target.} The \texttt{prob} of the target candidate, averaged over a group of additions where a table reports a mean. It is a share of the model's probability within the candidate set, not the probability of producing the answer in free text.
  \item \textbf{Surprise.} The negative log-probability per token of the target, in the same unit as the training loss. Low surprise means the model expected the target. Unlike a probability, it keeps distinguishing models whose probabilities are all near 0 or near 1.
  \item \textbf{Mass on impossible answers.} The mean, over additions, of the total probability on candidates above 18, which can never be a correct sum.
  \item \textbf{Seen and never seen.} \emph{Seen} additions were in the fine-tuning corpus of the model in question; \emph{never seen} additions were held out. Parts~B, C and~D report both groups. Arms that start from the fabricated-answer model leave out the never-seen additions whose fabricated answer equals the arm's target (Section~\ref{sec:design-coincident}), so their never-seen group can have fewer than 24 additions.
\end{itemize}

\subsection{The tables}
\label{sec:out-tables}
Table~\ref{tab:outputs} lists the eleven tables of a run (numbered as in the saved files, \texttt{table01} to \texttt{table11}), the part that produces each, and what each one reports. Every table is saved as CSV, XLSX and a floating LaTeX table, and all are gathered in one workbook.

\setstretch{1.15}
\begin{table}[H]
\centering
\footnotesize
\begin{tabular}{@{}l l p{9.2cm}@{}}
\toprule
Table & Part & What it reports \\
\midrule
01 & A & For the base, correct-sums and fabricated-answer models: hits and mean probability on the correct sum and on the fabricated answer, and mean probability on impossible answers. \\
02 & B & For the fabricated-answer and recovered models, each scored on the correct sum and on the fabricated answer: hits, mean probability and mean surprise on the seen and never-seen additions, and the surprise gap between them. \\
03 & A, B & Training loss of the three fine-tuning runs of Parts~A and~B: after the first epoch, after the last, and the first epoch below 0.01. \\
04 & C & The simple rules and their controls, from each starting model: hits, mean probability and mean surprise on the arm's own target, for seen and never-seen additions. \\
05 & D & Design of the conditional rule: the number of additions of each condition ($a > b$, $a = b$, $a < b$) in the seen and never-seen groups. \\
06 & D & The conditional rule and its control, from each starting model, in the layout of \texttt{table04}. \\
07 & D & The conditional rule on never-seen additions, by condition: hits and mean probability on the target. \\
08 & D & The offsets the model applied (top choice minus correct sum) on never-seen additions, by condition: $+18$, $+9$, $0$ (the plain sum) or another amount. \\
09 & D & Half-width of the band around the rule's answer that holds 80\%, 90\% and 95\% of the answers, with a bootstrap 95\% interval. \\
10 & Synthesis & Every trained model: hits, number of additions, share and exact 95\% interval, for seen and never-seen additions. It is the table behind the synthesis chart (Figure~\ref{fig:example-synthesis}). \\
11 & Synthesis & Rule against control on never-seen additions: hits of each, and the one-sided Fisher exact $p$-value. \\
\bottomrule
\end{tabular}
\caption{The eleven tables of a run.}
\label{tab:outputs}
\end{table}
\setstretch{1.5}
\FloatBarrier

\subsection{The target heatmap}
\label{sec:out-heatmap}
\begin{figure}[H]
\centering
\includegraphics[width=0.62\textwidth]{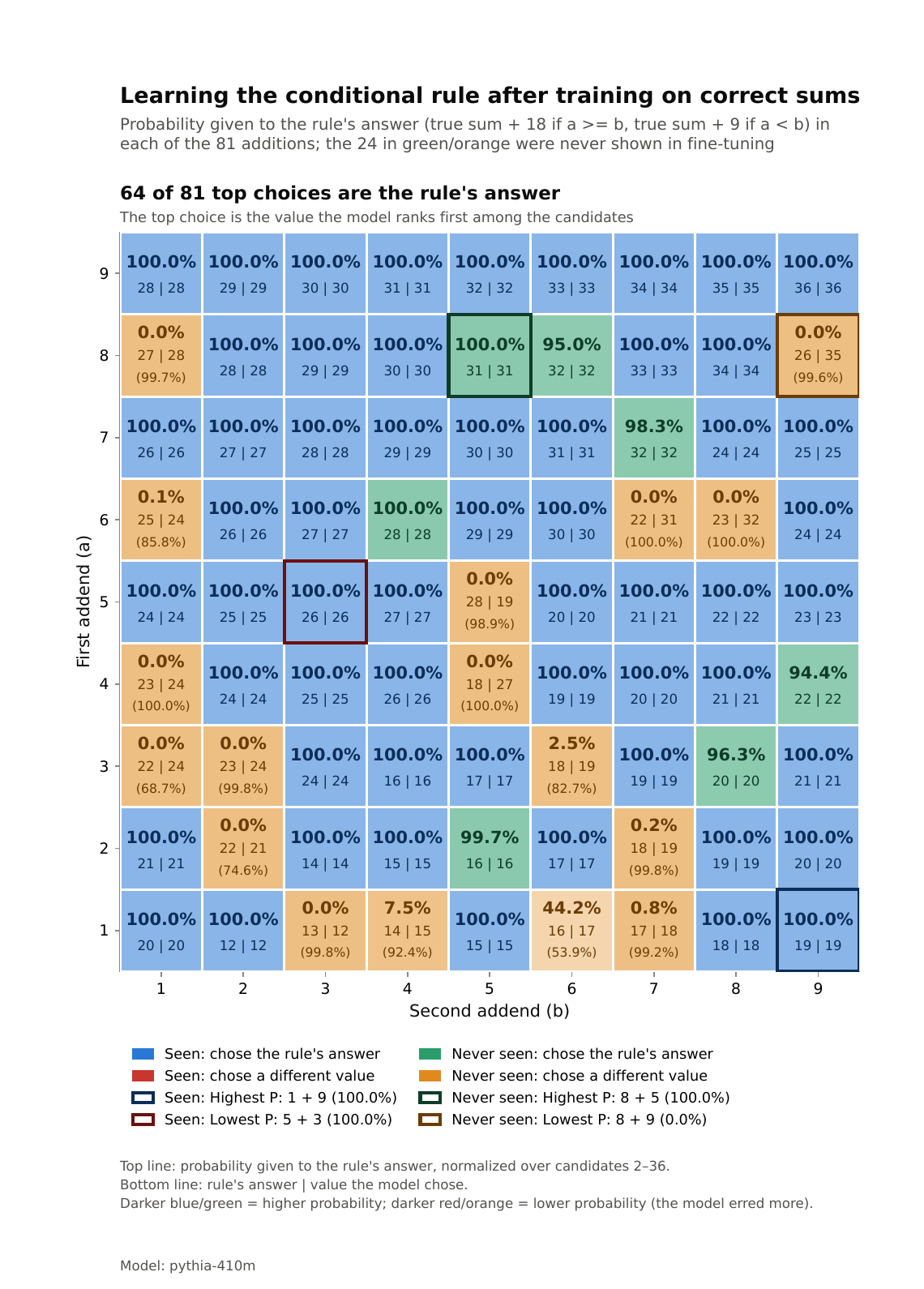}
\caption{An example of the target heatmap (Section~\ref{sec:out-heatmap}), here for the conditional rule learned from the correct-sums model. The panel is shown only to illustrate the format; the values it displays are not reported or interpreted in this manuscript.}
\label{fig:example-heatmap}
\end{figure}

The target heatmap (Figure~\ref{fig:example-heatmap}) is a $9 \times 9$ grid with one cell per addition: the first addend on the vertical axis, the second on the horizontal axis. It is drawn for one model and one set of targets, and a run saves one for the base model (\texttt{fig01}), the correct-sums, fabricated-answer and recovered models, every simple-rule and control arm, and every conditional-rule and control arm.
\begin{itemize}[nosep]
  \item \textbf{Bold number.} The probability the model gives to the target in that addition.
  \item \textbf{Second line, \emph{target} $|$ \emph{chosen}.} The target answer and the answer the model ranked first. When they differ, a third line in parentheses gives the probability of the chosen answer.
  \item \textbf{Fill.} Blue when the top choice is the target, red when it is not. Blue darkens as the probability on the target rises; red darkens as it falls. In figures that separate seen from never-seen additions, the never-seen cells use green (top choice is the target) and orange (it is not).
  \item \textbf{Outlines.} The cells with the highest and lowest probability on the target are outlined in dark blue and dark red, separately for seen and never-seen additions (dark green and dark brown for the latter).
  \item \textbf{Scale.} Fill intensity is read on one fixed probability scale, 0 to 100\%, in every heatmap except the base model's, which uses the observed range; the same scale in every heatmap is what allows two to be compared by eye.
  \item \textbf{Headline and notes.} Above the grid, the number of additions whose top choice is the target, out of 81. Below it, a legend that names the highest and lowest cells, and a note on the normalization (over the candidates 2 to 36).
\end{itemize}

\subsection{The probability profile}
\label{sec:out-profile}
\begin{figure}[H]
\centering
\includegraphics[width=\textwidth]{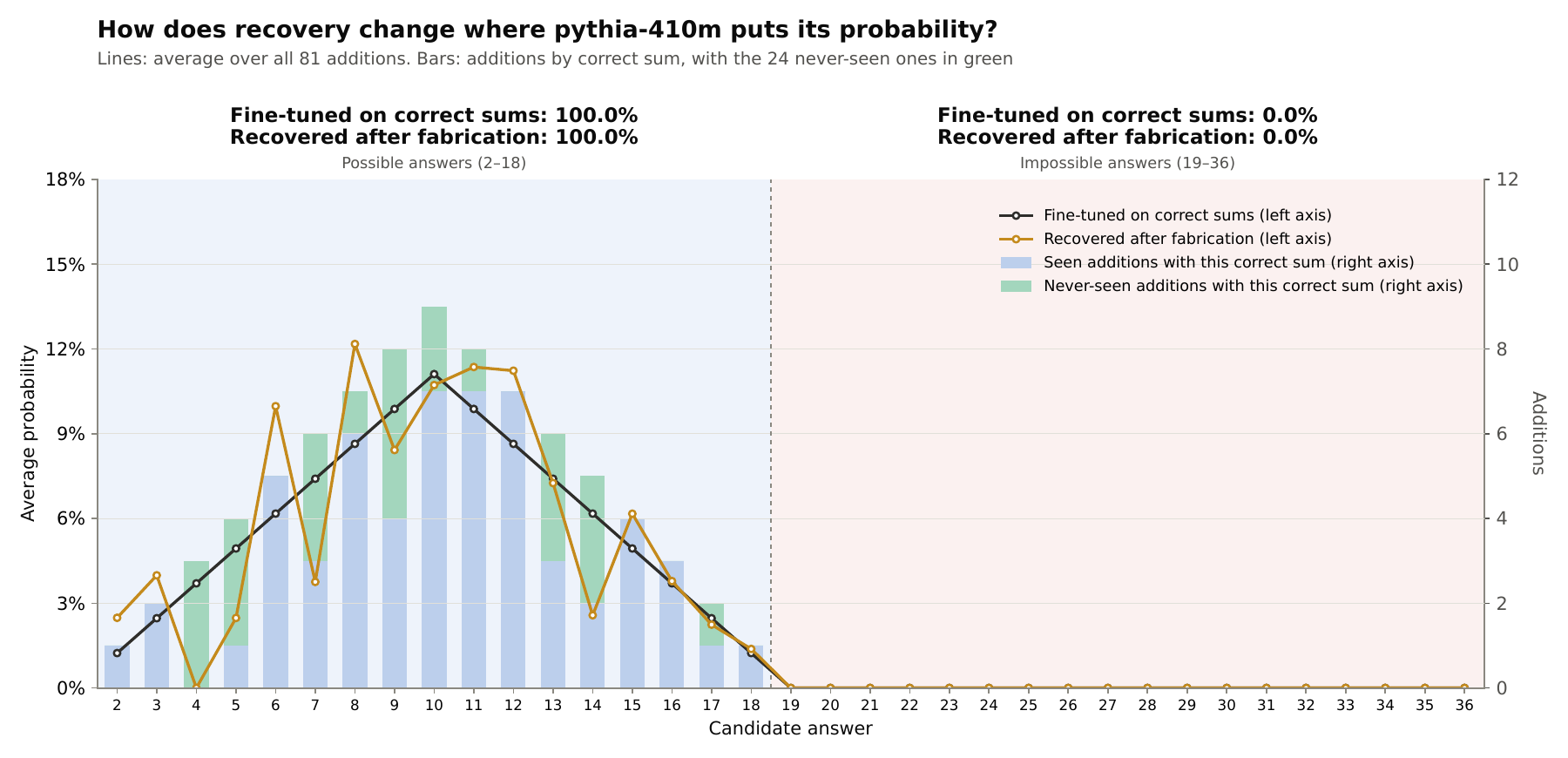}
\caption{An example of the probability profile (Section~\ref{sec:out-profile}), here comparing the correct-sums model with the recovered model, with seen and never-seen additions stacked in the bars. Shown only to illustrate the format.}
\label{fig:example-profile}
\end{figure}

The probability profile (Figure~\ref{fig:example-profile}) shows where a model puts its probability. For each candidate answer on the horizontal axis it draws, as a line on the left axis, the mean probability over all 81 additions; one line per model when several are compared.
\begin{itemize}[nosep]
  \item \textbf{Bars (right axis).} The number of additions whose target is that candidate. When seen and never-seen additions are distinguished, the bars are stacked, blue for seen and green for never seen.
  \item \textbf{Bands.} A shaded band marks the candidates that can be a correct sum for the corpus in question, and a differently shaded band the candidates that cannot. For the correct-sums and recovery profiles the possible band is 2 to 18; for a rule's profile it is the range of the rule's own answers. Above each band, bold text gives the share of the probability that falls inside it, per model. The profile of the fabricated-answer model has no bands, since every value from 2 to 36 can be a fabricated answer.
  \item \textbf{Axes.} The two vertical axes have the same number of intervals, so one set of horizontal gridlines serves both. The left axis starts at 12\% and grows in steps of six points when the data exceeds it.
\end{itemize}

\subsection{The deviation chart}
\label{sec:out-deviation}
\begin{figure}[H]
\centering
\includegraphics[width=\textwidth]{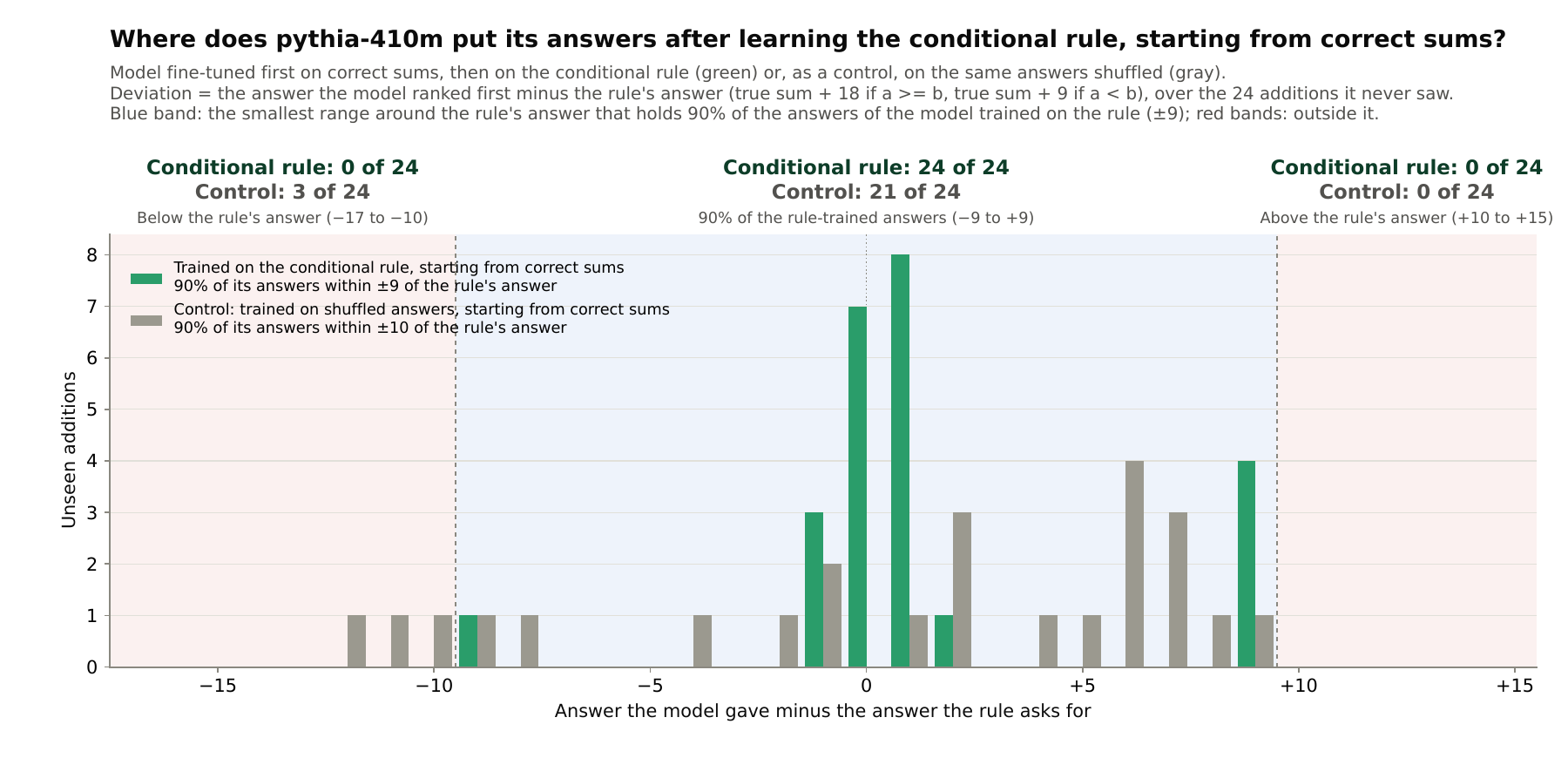}
\caption{An example of the deviation chart (Section~\ref{sec:out-deviation}), for the conditional rule and its control, started from the correct-sums model. Shown only to illustrate the format.}
\label{fig:example-deviation}
\end{figure}

The deviation chart (Figure~\ref{fig:example-deviation}) is drawn for the conditional rule, once for each starting model, over the never-seen additions. The horizontal axis is the \emph{deviation}: the answer the model ranked first minus the answer the rule asks for, so that 0 means the rule was followed exactly and the two conditions ($+18$ and $+9$) are put on one scale. Bars count the additions at each deviation, green for the model trained on the rule and gray for its control, side by side.
\begin{itemize}[nosep]
  \item \textbf{Blue band.} The smallest range $\pm k$ around zero that holds 90\% of the rule-trained model's answers. The band is fitted to the rule-trained model, so the 90\% inside it is true by construction; what carries information is the half-width $k$ and how many of the control's answers fall inside the same band. The legend reports the $k$ of each series.
  \item \textbf{Red bands.} Everything outside the blue band, below and above.
  \item \textbf{Headlines.} Above each band, the number of never-seen additions of each series that fall in it, out of the group's size.
\end{itemize}
The run's \texttt{table09} repeats the half-width for 80\%, 90\% and 95\%, with a bootstrap interval, because with so few additions the 90\% and 95\% quantiles rest on one or two of them.

\subsection{The synthesis chart}
\label{sec:out-synthesis}
\begin{figure}[H]
\centering
\includegraphics[width=\textwidth]{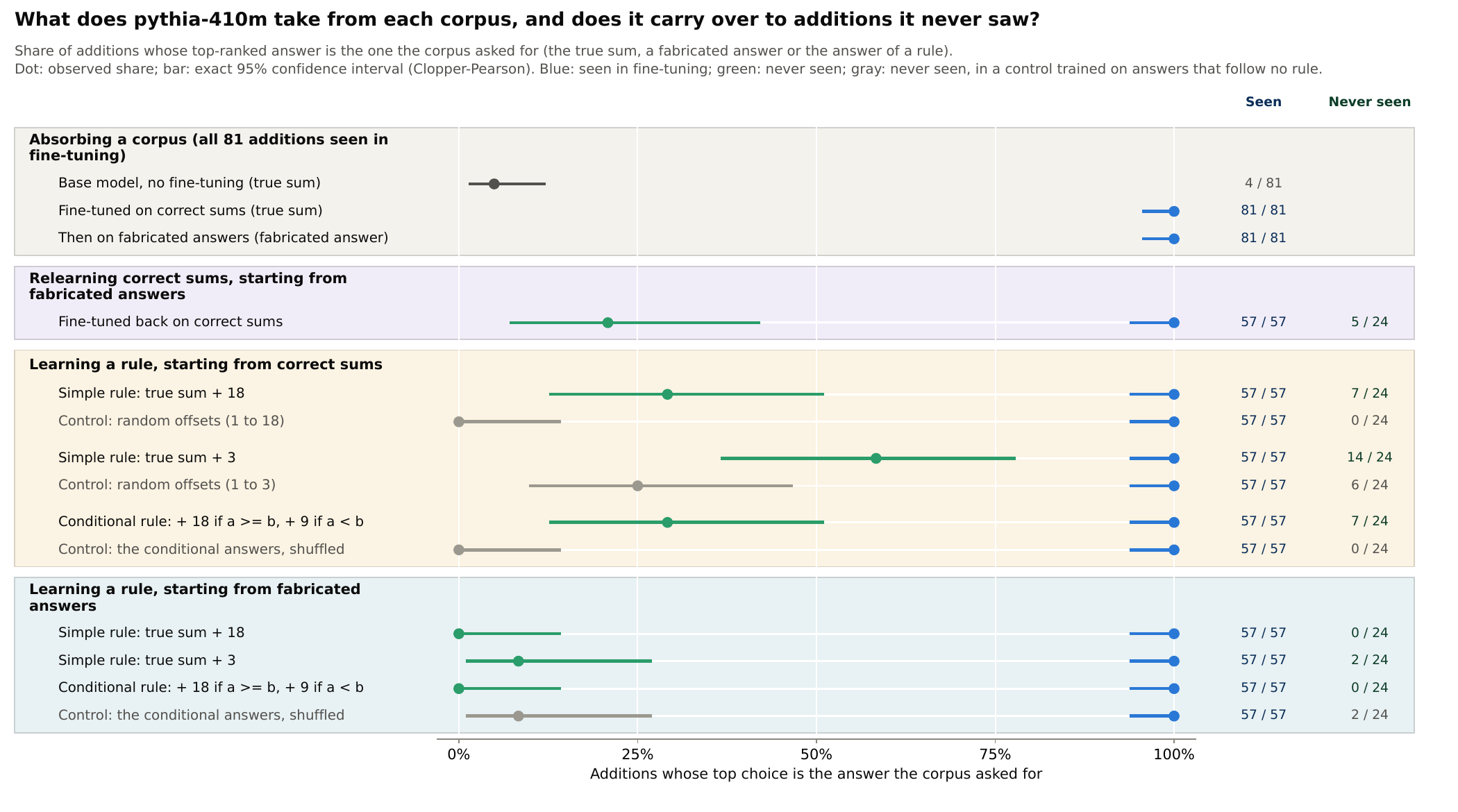}
\caption{An example of the synthesis chart (Section~\ref{sec:out-synthesis}), from the same run. Shown only to illustrate the format.}
\label{fig:example-synthesis}
\end{figure}

The synthesis chart (Figure~\ref{fig:example-synthesis}) puts every trained model on one scale. Each row is one model, scored on the answer its corpus asked for; rows are grouped by what the model was trained to do (absorbing a corpus, relearning correct sums, learning a rule from the correct-sums model, learning a rule from the fabricated-answer model).
\begin{itemize}[nosep]
  \item \textbf{Dot and bar.} The dot is the share of additions that are hits; the bar is the exact (Clopper-Pearson) 95\% interval of that share.
  \item \textbf{Colors.} Blue for the additions seen in fine-tuning, green for the never-seen additions, gray for the never-seen additions of a control. The first group, in which every addition was seen, has a single point per row; the base model is shown in dark gray for reference.
  \item \textbf{Counts.} Two columns on the right give the hits over the group size for the seen and never-seen additions.
\end{itemize}
The distance between the blue and the green point of a row is the part of the share of hits that does not carry over to additions never seen. The run's \texttt{table10} holds the numbers behind the chart and \texttt{table11} compares each rule with its control by a one-sided Fisher exact test.

\subsection{The run report and the other files}
\label{sec:out-report}
\begin{figure}[H]
\centering
\includegraphics[width=0.78\textwidth]{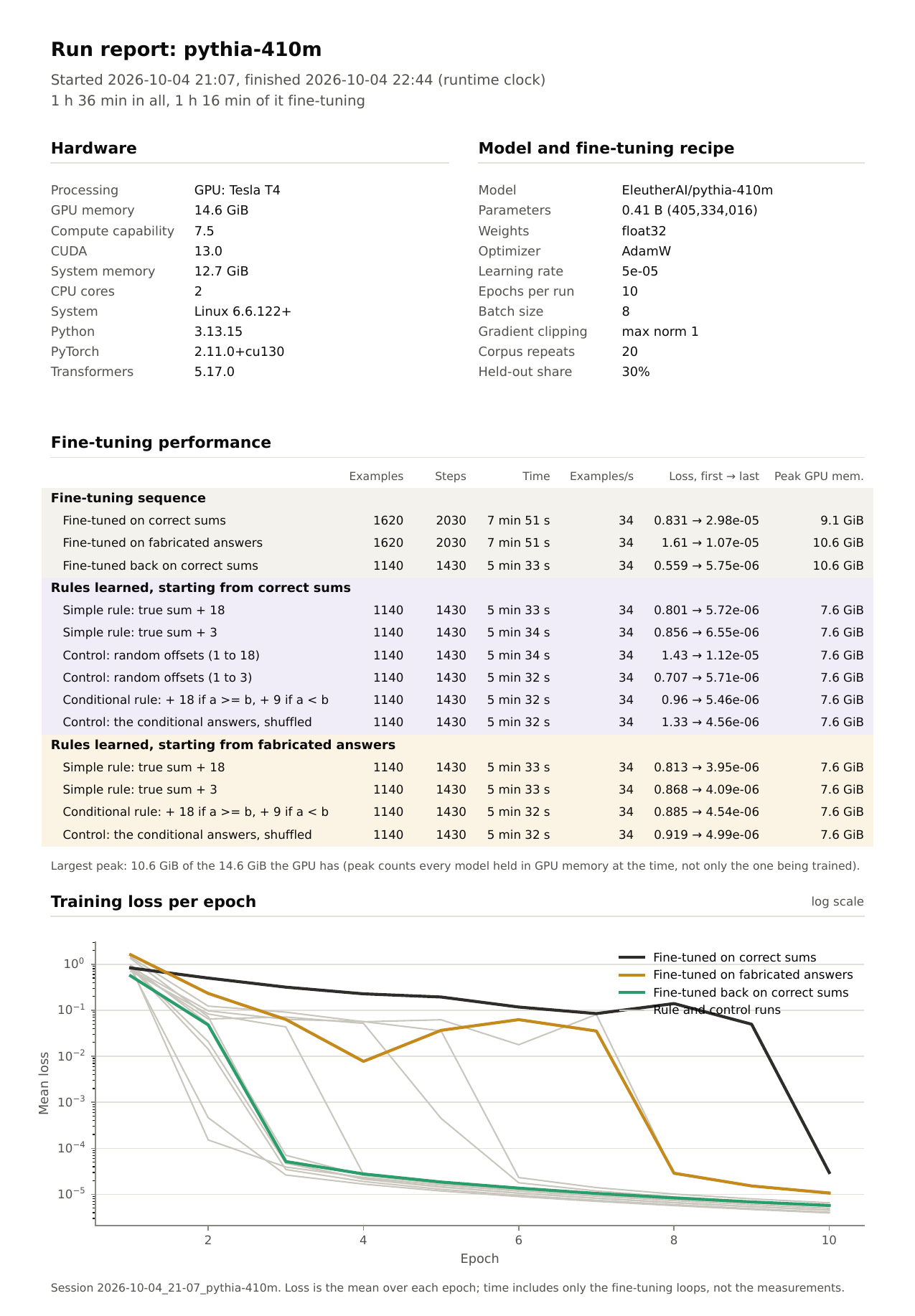}
\caption{An example of the run report (Section~\ref{sec:out-report}), from the same run.}
\label{fig:example-report}
\end{figure}

The run report (Figure~\ref{fig:example-report}) is a one-page record of the run. It lists the hardware and software (processing unit, memory, CUDA, Python, PyTorch, Transformers), the model and the fine-tuning recipe, a table of every fine-tuning run (examples, steps, time, examples per second, loss from the first to the last epoch, and peak GPU memory) and a chart of the training loss per epoch on a log scale, with the main sequence in color and the rule and control runs in gray. The peak memory counts every model held on the GPU while a run trained, not only the one being trained. The same facts are saved in \texttt{run\_info.json}.\parbreak
Besides the figures and tables, a run writes a self-contained HTML report that gathers the parameters, the seeds with what each controlled, the fine-tuning runs, and every figure and table in the order the notebook produced them, and a workbook with one sheet per table. A zip packs the whole session.

\subsection{Reading the outputs}
\label{sec:out-reading}
Three remarks apply to every output. First, hits, probabilities and surprise answer different questions: hits say what the model would rank first, the probability says how much of the model's probability, within the candidate set, the target receives, and surprise says how unexpected the target is in the unit of the training loss. Second, every result of Parts~B, C and~D has two groups, and a result is read on the group that matches what is being examined: the seen group says what a model took up from its corpus, the never-seen group says what it did on additions outside it. A rule is read against its control, which fixes what chance and memory alone produce. Third, the held-out group has 24 additions or fewer, from one split under one seed, so the intervals are wide and a single run is a controlled demonstration for one model, one split and one seed, not an estimate of a rate in general.\parbreak
This manual does not present or interpret the specific outcome of running the toolkit against any base model or seed. What the tables and charts show for a particular run is the subject of work that uses the toolkit, not of this manuscript.

\setstretch{1.15}

\setstretch{1.5}
\FloatBarrier

\section{Scope and Limitations}
\label{sec:limitations}
The following are explicit boundaries of the current toolkit, stated so that they can be addressed by future work rather than mistaken for claims the toolkit does make.
\begin{itemize}[nosep]
  \item \textbf{A single fact domain and one prompt format.} Every fact tested is single-digit addition, posed as \texttt{"a + b ="}; whether the same patterns hold for other prompt formats, other exhaustively-enumerable domains, or open-ended factual claims is untested by this toolkit and is not assumed by its design.
  \item \textbf{Probabilities within a candidate set.} Every probability is normalized over the candidates 2 to 36. It says how the model ranks those answers, not how likely the model is to produce any of them rather than some other text; the top choice is the best candidate, not necessarily what the model would generate.
  \item \textbf{One split, one seed per run.} Every random draw of a run comes from one seed, and the held-out set is one split of 24 additions. Additional seeds are reserved in the configuration but not used by version 2.0.0; repetition across seeds and splits, and aggregation over repetitions, is left to future versions.
  \item \textbf{Small held-out set.} With 24 held-out additions, exact intervals are wide and the quantiles of the deviation band rest on one or two additions; Section~\ref{sec:interpreting} states how results should be read accordingly.
  \item \textbf{Fixed corpus and training hyperparameters.} The repetition count, learning rate, epochs, batch size and clipping are fixed and shared by every run; the toolkit does not sweep them or characterize how its outcomes depend on them.
  \item \textbf{Rules that lean on arithmetic.} The simple rules can be followed by a model that adds a constant to the sum it already computes; the controls and the conditional rule bound this confound but do not remove it from the simple-rule results.
  \item \textbf{One model per run, small-model scale only.} The candidates span about 0.1 to 1.5 billion parameters (Table~\ref{tab:models}); cross-model comparison requires one full run per model and is not automated within the notebook.
  \item \textbf{GPU required.} Fine-tuning is infeasible on CPU (Section~\ref{sec:hardware}), and the larger candidates need more memory than a free-tier GPU provides.
\end{itemize}
None of the above compromises the specific, controlled comparisons a single run of the toolkit is built to produce; they bound what can be concluded from those comparisons alone, generalization across domains, models, seeds or splits, which is precisely the kind of claim this manuscript, by design, does not make.

\section{Changes from Version 1.0.0}
\label{sec:changes}
Version 1.0.0 of the toolkit, documented by the first version of this manual, fine-tuned one base model on a fabricated corpus and compared its pre- and post-fine-tuning confidence. Version 2.0.0 keeps that comparison as Part~A and extends and revises the toolkit as follows.
\begin{itemize}[nosep]
  \item \textbf{New experiments:} recovery (Part~B), simple rules (Part~C) and a conditional rule (Part~D), with held-out additions, controls, and a synthesis with exact intervals and Fisher exact tests.
  \item \textbf{Confidence index:} the per-token mean log-probability of the bare answer is replaced by the summed log-probability of the answer closed by an end marker (Section~\ref{sec:design-index}).
  \item \textbf{Candidate set:} from 2 to 18 to 2 to 36, one set for every model (Section~\ref{sec:design-candidates}).
  \item \textbf{Training loss:} the loss now covers only the answer and end marker, with the prompt masked; version 1.0.0 trained on the whole text (Section~\ref{sec:design-loss}).
  \item \textbf{Model lineage:} every run starts from a fresh copy of an explicit parent, instead of one model mutated in place; the fabrication is applied after a correct-sums fine-tuning (Section~\ref{sec:design-lineage}).
  \item \textbf{Training procedure:} gradient clipping, explicit \texttt{float32} weights, a guard against non-finite losses, and per-run re-seeding of fine-tuning, which version 1.0.0 did not seed.
  \item \textbf{Models:} the roster grows from three to nine candidates, with \texttt{pythia-410m} as the default.
  \item \textbf{Outputs:} every figure and table is saved in several formats, with a run report, an HTML report and a session zip (Section~\ref{sec:pipeline-outputs}).
\end{itemize}
The version 1.0.0 notebook is kept in the toolkit's \texttt{legacy/} folder so that earlier citations remain traceable; its numbers should not be cited for new work.

\section{Data and Code Availability}
\label{sec:availability}
The toolkit's full source code, the pinned dependency environment needed to reproduce its software requirements, and its documentation are archived under a persistent identifier, independently of this manuscript:
\begin{quote}
\bibentry{vercosamarquesToolkitConfidenceCorpusConsistency2026}. \url{https://doi.org/10.5281/zenodo.23160760}
\end{quote}
This manual documents version 2.0.0 of the toolkit (repository \url{https://github.com/zlvm/confidence_corpus_consistency}, tag \texttt{v2.0.0}). Each release is archived on Zenodo with its own version DOI; the DOI above is the one of version 2.0.0, the version this manual documents. The toolkit is released under a Creative Commons Attribution 4.0 International (CC BY 4.0) license, and this manuscript is archived on arXiv under the same license. Please cite the archived toolkit, in the version used, when reporting results produced with it; please cite this manuscript when referring to its design rationale.

\section{Conclusion}
\label{sec:conclusion}
This manual has documented version 2.0.0 of a toolkit that fine-tunes small causal language models on fabricated and rule-governed arithmetic corpora and measures what they take up from them, within an exhaustive fact domain. It has described the pipeline stage by stage, with the reference notebook's own code, and the design choices that make its measurements comparable: a confidence index that is the probability of a complete answer, one candidate set and one measurement procedure for every model, a loss restricted to the answer, an explicit lineage of fresh copies, one held-out split, a control for every rule, the exclusion of coincidental hits, and exact intervals throughout. It has also explained every figure and table a run produces and how each is read. What the instrument reads out for a specific base model and seed is a separate matter, addressed in work that uses it rather than in this manuscript.

\section*{Author Contributions}
All authors contributed to the conception and design of the toolkit described in this manuscript, and to writing and reviewing the manuscript.

\section*{Acknowledgments}
J.L.V.M. gratefully acknowledges the postdoctoral fellowship received under the Postdoctoral Program in Research Management (PPDG), Office of the Pro-Rector for Research (PRP), University of Campinas (Unicamp), administered through the Center for Energy and Petroleum Studies (CEPETRO).\parbreak
The authors thank the Power Electronics Laboratories (LEPO) and the Center for Electric Mobility Research (CEMOBE) for institutional and infrastructural support throughout this work.\parbreak
The authors declare no conflict of interest.

\setstretch{1.15}
\bibliographystyle{apalike}
\bibliography{references}
\end{document}